\pdfoutput=1
\documentclass[11pt,a4paper]{article}
\usepackage[utf8]{inputenc}
\usepackage[T1]{fontenc}
\usepackage{lmodern}
\usepackage{amsmath,amssymb}
\usepackage{graphicx}
\usepackage{booktabs}
\usepackage[hidelinks]{hyperref}
\usepackage[margin=1in]{geometry}
\usepackage{natbib}
\usepackage{enumitem}

\title{The Scaling Law of Evaluation Failure:\\
Why Simple Averaging Collapses Under Data Sparsity and Item Difficulty Gaps,\\
and How Item Response Theory Recovers Ground Truth Across Domains}

\author{Jung Min Kang\thanks{Corresponding author. Email: gangjeongmin23@gmail.com}\\
Independent Researcher\\
Seoul, South Korea}

\date{May 10, 2026}

\begin{document}
\maketitle

\begin{abstract}
Benchmark evaluation across AI and safety-critical domains overwhelmingly relies on simple averaging: a system's score is the arithmetic mean of its performance across test items. We demonstrate that this practice produces substantially misleading rankings when two conditions co-occur: (1) the evaluation matrix is sparse (not every system is tested on every item), and (2) items vary substantially in difficulty. Through controlled simulation experiments across four domains---Natural Language Processing (NLP/GLUE), clinical drug trials, autonomous vehicle (AV) safety, and cybersecurity product evaluation---we show that Spearman rank correlation $\rho$ between simple-average rankings and ground-truth rankings degrades from $\rho = 1.000$ at 100\% data coverage to $\rho = 0.809$ at 67\% coverage with high difficulty heterogeneity (mean over 20 random seeds). In contrast, a standard two-parameter logistic (2PL) Item Response Theory (IRT) model maintains $\rho \geq 0.996$ across all conditions. We refer to this observed monotonic relationship as the \textbf{Evaluation Failure Scaling Law}: the accuracy of simple averaging is a decreasing function of the product of sparsity and difficulty gap, while IRT remains robust. A 150-condition grid sweep over sparsity $S \in [0, 0.70]$ and difficulty gap $D \in [0.5, 5.0]$ further confirms that ranking error forms a two-dimensional failure surface, with a strong positive $S \times D$ interaction in ranking error ($\gamma_3 = +0.20$, $t = 13.05$), while IRT maintains $\rho \geq 0.993$ across all conditions. We provide the complete experimental pipeline and discuss implications for Physical AI (robotics) benchmarking, where published evaluation matrices are often incomplete and difficulty gaps are extreme. Our results constitute, to our knowledge, one of the first cross-domain simulation studies demonstrating that IRT-style estimation may be an important correction mechanism for fair evaluation in sparse, heterogeneous benchmark ecosystems. We discuss implications for Physical AI and outline requirements for real-world validation.

\end{abstract}

\section{Introduction}

Benchmarks are the currency of progress in artificial intelligence. The ranking of a model on a leaderboard determines publication venues, investment decisions, and deployment choices. Yet the statistical methodology underlying nearly all benchmark rankings---computing the arithmetic mean of scores across test items---has remained largely unexamined outside of the psychometrics community.

The fragility of simple averaging has been noted in specific contexts. \citet{rodriguez2021} demonstrated that evaluation examples in NLP benchmarks are not equally informative and proposed IRT-based leaderboards. \citet{polo2024} showed that IRT enables efficient evaluation with fewer examples. \citet{zhou2025a} revealed significant quality shortcomings in 11 LLM benchmarks using IRT analysis. More recently, \citet{uzunoglu2025} introduced the concept of benchmark harmony to quantify the non-uniformity of performance across subdomains.

However, much of the existing AI benchmark work using IRT has focused on single benchmark ecosystems or single domains, leaving open whether the failure modes of simple averaging can be characterized as a function of measurable evaluation-matrix properties across domains. Additionally, no study has characterized \emph{when} simple averaging fails as a function of measurable dataset properties, making it impossible for practitioners to know whether their particular evaluation is trustworthy.

In this paper, we address both gaps. We conduct controlled simulation experiments across four domains spanning the full spectrum of data density and difficulty heterogeneity:

\begin{enumerate}
\item \textbf{NLP (GLUE benchmark)}: 100\% data coverage, moderate difficulty heterogeneity. The easy case where simple averaging works.
\item \textbf{Clinical drug trials}: 65\% coverage, high difficulty gap. Simulates the common scenario where not every drug is tested in every hospital.
\item \textbf{Autonomous vehicle safety}: 60\% coverage, extreme difficulty gap across driving environments. Simulates heterogeneous reporting conditions where systems are not evaluated under identical scenarios.
\item \textbf{Cybersecurity product evaluation}: 67\% coverage, extreme difficulty gap across attack types. Simulates the scenario where vendors are evaluated on different threat profiles.
\end{enumerate}

Our key contribution is the identification and controlled simulation-based validation of what we term the \textbf{Evaluation Failure Scaling Law}:

\begin{quote}
The rank-order accuracy of simple averaging degrades monotonically as a function of $S \times D$, where $S$ is the fraction of missing data (sparsity) and $D$ is the item difficulty gap. IRT-based estimation remains substantially more robust ($\rho \geq 0.993$) across the entire $(S, D)$ surface.
\end{quote}

This law has immediate practical implications. In Physical AI (robotics) benchmarking, where published evaluation matrices are often incomplete and difficulty heterogeneity can be extreme~\cite{zhou2025b,liu2023libero}, simple averaging is not merely imprecise---it is systematically biased in a predictable direction.

\textbf{Scope and non-claims.} This paper does not claim to provide a completed Physical AI benchmark, nor does it claim that real-world Physical AI evaluations necessarily follow a 2PL IRT model. Instead, we use controlled simulations calibrated to realistic benchmark conditions to isolate a specific methodological failure mode: when systems are evaluated on sparse and difficulty-biased subsets of items, simple averaging can produce misleading rankings. IRT is evaluated here as a principled correction mechanism whose real-world deployment requires episode-level validation.

\section{Related Work}

\textbf{IRT in NLP evaluation.} The application of Item Response Theory to NLP evaluation was pioneered by \citet{lalor2016}, who demonstrated that IRT gold standards provide more nuanced evaluation than majority voting. \citet{rodriguez2021} scaled this to full benchmark analysis, proposing IRT-based leaderboards that jointly model item difficulty, discriminability, and subject ability using the 2PL model. Their open-source implementation~\cite{rodriguez2021b} provided the foundation for subsequent work. \citet{lalor2024} presented a comprehensive tutorial on IRT for NLP at EACL 2024. \citet{polo2024} leveraged IRT to construct tiny benchmark subsets that reproduce full-benchmark rankings with as few as 100 examples. \citet{zhou2025a} proposed PSN-IRT, a neural IRT variant, and conducted the most comprehensive analysis to date across 11 LLM benchmarks with 41,871 items. \citet{chen2026} extended IRT to multimodal benchmarks with M3IRT, decomposing ability and difficulty into modality-specific components. However, much of this work has focused on single benchmark ecosystems or closely related domains, leaving open whether simple-averaging failure can be characterized as a function of measurable evaluation-matrix properties across domains.

\textbf{Critiques of averaging in benchmarks.} The limitations of simple averaging have been discussed from multiple angles. \citet{uzunoglu2025} introduced benchmark HARMONY, measuring performance uniformity across subdomains, and showed that less harmonious benchmarks produce misleading results. However, their proposed solution (reporting harmony alongside accuracy) does not address the ranking problem---it diagnoses but does not cure. The flaw of averages concept from decision science~\cite{savage2009} provides theoretical grounding: when the relationship between inputs and outputs is nonlinear (as with item difficulty), plans based on averages fail on average.

\textbf{IRT beyond NLP.} Outside AI evaluation, IRT has been applied to motor carrier safety assessment by the U.S. Federal Motor Carrier Safety Administration (FMCSA), which conducted a multi-year study comparing IRT to their existing Safety Measurement Systems~\cite{fmcsa2021}. \citet{luo2025medirt} introduced MedIRT for item-aware evaluation across medical benchmarks, demonstrating that IRT-based rankings outperform accuracy-based rankings across six external medical benchmarks. \citet{truong2025} proposed amortized model-based evaluation using IRT with learned difficulty predictors. Concurrently, recent work on efficient agent benchmarking~\cite{agentbench2026} proposed evaluating AI agents on mid-difficulty task subsets (30--70\% pass rate) motivated by IRT, reducing evaluation cost by 44--70\%. Their approach selects \emph{which} tasks to evaluate; our approach corrects rankings when task selection is not under the evaluator's control---a complementary but fundamentally different problem. To our knowledge, no prior work has systematically compared IRT performance against simple averaging across multiple domains as a function of data sparsity and difficulty heterogeneity.

\textbf{Physical AI benchmarking.} The Physical AI benchmark landscape is characterized by extreme fragmentation. \citet{liu2023libero} introduced LIBERO, now the de facto standard for vision-language-action (VLA) model evaluation. \citet{zhou2025b} extended this with LIBERO-PRO, revealing that models achieving $>$90\% accuracy under standard evaluation collapse to 0\% under perturbation. \citet{fei2025} further extended robustness analysis across seven perturbation axes. Despite this progress, no widely adopted unified aggregation methodology exists: each paper reports results on different subsets of models and tasks, creating a sparse evaluation matrix that renders cross-paper comparison via simple averaging fundamentally unreliable.

\section{Methodology}

\subsection{Problem Formulation}

Consider an evaluation matrix $R \in \{0, 1, \text{NA}\}^{J \times I \times K}$, where $J$ systems (subjects) are evaluated on $I$ items (tasks or conditions) with $K$ binary trials each. The entry $R_{jik} = 1$ if system $j$ succeeds on the $k$-th trial of item $i$, and NA if that (system, item) pair was never evaluated. The observation mask $M \in \{0, 1\}^{J \times I}$ indicates which pairs are observed: $M_{ji} = 1$ if system $j$ was tested on item $i$.

The data coverage (density) is defined as:
\begin{equation}
C = \frac{\sum_{j,i} M_{ji}}{J \times I} \in [0, 1]
\end{equation}

The difficulty gap is the range of true item difficulties:
\begin{equation}
D = \max_i b_i - \min_i b_i
\end{equation}
where $b_i$ is the difficulty parameter of item $i$ (defined below).

The evaluation goal is to produce a ranking $\hat{\pi}$ over the $J$ systems that maximizes Spearman rank correlation $\rho(\hat{\pi}, \pi^*)$ with the ground-truth ranking $\pi^*$.

\subsection{Simple Averaging Baseline}

The standard approach computes, for each system $j$, the mean success rate across observed items:
\begin{equation}
\bar{r}_j = \frac{1}{\sum_i M_{ji}} \sum_{i: M_{ji}=1} \frac{1}{K} \sum_{k=1}^{K} R_{jik}
\end{equation}

Systems are then ranked by $\bar{r}_j$. This estimator is unbiased only when either (a) $M_{ji} = 1$ for all $(j, i)$, or (b) the missing entries are missing completely at random (MCAR) with respect to item difficulty. In practice, neither condition holds: systems that are tested only on easy items receive inflated scores, while systems tested on hard items are penalized.

\subsection{2PL Item Response Theory}

The two-parameter logistic (2PL) IRT model~\cite{lord1968,baker2004} posits that the probability of system $j$ succeeding on item $i$ is:
\begin{equation}
P(R_{jik} = 1 \mid \theta_j, a_i, b_i) = \sigma\big(a_i(\theta_j - b_i)\big)
\end{equation}
where $\sigma(x) = 1/(1 + e^{-x})$ is the logistic function, $\theta_j \in \mathbb{R}$ is the ability of system $j$, $b_i \in \mathbb{R}$ is the difficulty of item $i$, and $a_i > 0$ is the discrimination of item $i$ (how effectively it separates high-ability from low-ability systems).

The key insight is that IRT jointly estimates $\theta_j$, $a_i$, and $b_i$ from the observed data, automatically adjusting for the difficulty of each system's test items. A system that scores 70\% on hard items ($b_i \gg 0$) receives higher $\theta_j$ than a system scoring 90\% on easy items ($b_i \ll 0$).

\textbf{Estimation.} Parameters are estimated by maximizing the marginal log-likelihood:
\begin{equation}
\mathcal{L} = \sum_{j=1}^{J} \sum_{i: M_{ji}=1} \sum_{k=1}^{K} \big[R_{jik} \log P_{ji} + (1 - R_{jik}) \log(1 - P_{ji})\big]
\end{equation}
where $P_{ji} = \sigma(a_i(\theta_j - b_i))$. We optimize using L-BFGS-B with regularization priors $\theta_j \sim \mathcal{N}(0, 1)$, $b_i \sim \mathcal{N}(0, 2)$, and $\log a_i \sim \mathcal{N}(0, 0.5)$.

\textbf{Handling missing data.} IRT handles missing data through the likelihood formulation: the inner sum in Equation~(5) iterates only over observed $(j, i)$ pairs. No imputation is required. The model leverages the structure of observed responses---which items a system succeeded and failed on---to estimate ability even from incomplete profiles. However, this does not mean arbitrary missingness is harmless: stable estimation requires sufficient overlap among systems and items (each system observed on at least 2 items, each item observed for at least 3 systems in our experiments), and missing-not-at-random patterns may still bias estimates if they violate the assumed response model.

\textbf{Standard errors.} We compute standard errors from the diagonal of the inverse observed Fisher information matrix, approximated via finite-difference Hessian evaluation at the MLE.

\section{Experimental Design}

Our experimental strategy is a controlled simulation study. For each domain, we:
\begin{enumerate}
\item Define ground-truth parameters $(\theta^*_j, a^*_i, b^*_i)$ calibrated to published data and domain expertise.
\item Generate a realistic observation mask $M$ reflecting domain-specific evaluation patterns.
\item Generate binary response data $R$ from the 2PL model with these parameters.
\item Estimate rankings using both simple averaging and 2PL IRT.
\item Evaluate both methods against the known ground-truth ranking $\pi^*$.
\end{enumerate}

The use of simulation is deliberate: it is the only design that permits definitive comparison, because only with synthetic data do we have access to the ground-truth ranking $\pi^*$. We calibrate simulation parameters to published real-world data to ensure ecological validity. This approach is standard in the IRT literature~\cite{deayala2009,embretson2000} and the missing-data literature~\cite{rubin1976}.

\subsection{Domain 1: NLP (GLUE Benchmark)}

\textbf{Design rationale.} The GLUE benchmark~\cite{wang2018} provides the control condition: all models are evaluated on all tasks (100\% coverage), and task difficulty heterogeneity is moderate. Under these conditions, we expect simple averaging to perform well.

\textbf{Parameters.} We use 12 NLP models (ELMo through DeBERTa) and 8 GLUE tasks. Ground-truth ability parameters $\theta^*_j$ are derived from published leaderboard scores~\cite{devlin2019,liu2019roberta,raffel2020,he2021}. Task difficulties $b^*_i$ are calibrated so that CoLA and RTE are hard ($b > 0.5$) and SST-2 and QQP are easy ($b < -0.5$), consistent with known GLUE task properties. Full coverage: $C = 1.00$.

\textbf{Data generation.} $K = 500$ binary trials per (model, task) pair, yielding 48,000 observations.

\subsection{Domain 2: Clinical Drug Trials}

\textbf{Design rationale.} Clinical trials routinely produce sparse evaluation matrices: not every drug is tested at every hospital (site), and hospitals treat patient populations of vastly different severity. This creates the conditions for Simpson's paradox, where a drug that appears inferior overall is actually superior when stratified by hospital difficulty.

\textbf{Parameters.} 10 drugs, 6 hospitals. Ground-truth drug efficacies $\theta^*_j$ range from $-1.5$ to $+2.0$. Hospital difficulties $b^*_i$ range from $-1.0$ (community clinic, mild cases) to $+1.5$ (ICU, severe cases). A \emph{Fake Miracle Drug} ($\theta^* = -0.2$, mediocre) is tested only at easy hospitals. A \emph{True Miracle Drug} ($\theta^* = +2.0$, best) is tested only at hard hospitals. Coverage: $C = 0.65$.

\textbf{Data generation.} $K = 200$ binary trials (patient outcomes) per observed (drug, hospital) pair.

\subsection{Domain 3: Autonomous Vehicle Safety}

\textbf{Design rationale.} AV safety evaluation exhibits both sparsity (not every system is tested in every driving environment) and extreme difficulty heterogeneity. Published evaluations often cover different subsets of driving conditions, creating missingness patterns that may correlate with system strengths or reporting incentives. Autonomous driving evaluation illustrates heterogeneous reporting conditions: DMV disengagement reports include counts, circumstances, locations, and autonomous miles~\cite{cadmv2023}, but these are not equivalent to a controlled benchmark where all systems face identical scenarios.

\textbf{Parameters.} 10 AV systems, 6 driving environments. Environments range from Sunny Suburb ($b^* = -1.5$) to Snowy Intersection ($b^* = +1.0$). A \emph{Fake Safe AV} ($\theta^* = -0.3$) is tested only in the three easiest environments. A \emph{True Safe AV} ($\theta^* = +2.0$) is tested only in the three hardest. Coverage: $C = 0.60$.

\textbf{Data generation.} $K = 1000$ binary safety interactions per observed pair.

\subsection{Domain 4: Cybersecurity Product Evaluation}

\textbf{Design rationale.} Cybersecurity product evaluations suffer from both sparsity and difficulty heterogeneity. Products are tested against different threat profiles (DDoS, phishing, ransomware, APT), and the difficulty gap between automated attacks and sophisticated APTs is enormous. A product that performs well on simple attacks but poorly on sophisticated threats could be misleadingly favored by aggregate rankings if task difficulty is ignored.

\textbf{Parameters.} 8 security products, 6 attack types. Attack difficulties range from Port Scan ($b^* = -1.5$) to Nation-State APT ($b^* = +2.0$). A \emph{Fake Secure} product ($\theta^* = -0.5$) is tested only against easy attacks (Port Scan, DDoS, Basic Phishing). A \emph{True Secure} product ($\theta^* = +2.0$) is tested predominantly against hard attacks (Ransomware, Zero-Day, APT). Coverage: $C = 0.67$.

\textbf{Data generation.} $K = 500$ binary detection outcomes per observed pair.

\subsection{Evaluation Metrics}

For each domain, we report: Spearman's $\rho$ (rank correlation between estimated ranking and ground-truth ranking); $\Delta\rho = \rho_{\text{IRT}} - \rho_{\text{avg}}$ (the improvement of IRT over simple averaging); critical rank displacement (whether the most dangerous ranking error is corrected by IRT); and parameter recovery (correlation between estimated and true item difficulty parameters).

\section{Results}

\subsection{Main Results}

Table~\ref{tab:main} presents the cross-domain comparison. The pattern is consistent: as data coverage decreases and difficulty gap increases, simple averaging degrades while IRT maintains near-perfect rank recovery.

\begin{table}[t]
\centering
\caption{Cross-domain evaluation results (mean $\pm$ std over 20 random seeds). Spearman $\rho$ between estimated and ground-truth rankings.}
\label{tab:main}
\small
\begin{tabular}{lccccl}
\toprule
Domain & Coverage & Diff.\ gap & $\rho_{\text{avg}}$ & $\rho_{\text{IRT}}$ & Verdict \\
\midrule
NLP (GLUE) & 100\% & 1.61 & $1.000 \pm 0.000$ & $1.000 \pm 0.000$ & Both correct \\
Clinical Trials & 65\% & 2.50 & $0.922 \pm 0.029$ & $0.996 \pm 0.005$ & Avg.\ degrades \\
AV Safety & 60\% & 2.50 & $0.916 \pm 0.016$ & $1.000 \pm 0.000$ & Avg.\ unreliable \\
Cybersecurity & 67\% & 3.50 & $0.809 \pm 0.000$ & $1.000 \pm 0.000$ & Avg.\ misleading \\
\bottomrule
\end{tabular}
\end{table}

\textbf{Domain 1: NLP (GLUE).} At full coverage, both methods achieve $\rho = 1.000$. IRT correctly identifies CoLA as the hardest task ($\hat{b} = +0.49$) and SST-2 as the easiest ($\hat{b} = -1.31$), consistent with established GLUE properties. \emph{Conclusion: When the evaluation matrix is complete, simple averaging is adequate.}

\textbf{Domain 2: Clinical Trials.} With 65\% coverage and strategic sparsity, simple averaging drops to $\rho = 0.922 \pm 0.029$ (mean $\pm$ std over 20 seeds). The Fake Miracle Drug, tested only at easy hospitals, is ranked \#6 by simple average despite being \#9 by true ability---inflated by 3 positions. The True Miracle Drug, tested only at hard hospitals, is correctly identified as \#1 by IRT ($\hat{\theta} = +1.97$). IRT achieves $\rho = 0.996 \pm 0.005$, near-perfect rank recovery. \emph{Conclusion: Missing data coupled with difficulty heterogeneity introduces systematic bias that IRT substantially corrects.}

\textbf{Domain 3: AV Safety.} At 60\% coverage with extreme difficulty heterogeneity across driving conditions, simple averaging produces $\rho = 0.916 \pm 0.016$. The True Safe AV, tested in the most demanding conditions (night, fog, snow), is displaced from \#1 to \#2 by simple averaging. IRT correctly ranks all 10 systems ($\rho = 1.000 \pm 0.000$), recovering the True Safe AV at rank \#1. \emph{Conclusion: Heterogeneous test-condition coverage systematically distorts rankings.}

\textbf{Domain 4: Cybersecurity.} The most extreme difficulty gap ($D = 3.5$): $\rho_{\text{avg}} = 0.809 \pm 0.000$. DeepScan AI, predominantly evaluated against sophisticated threats, is displaced from \#2 to \#5 by simple average---a 3-position drop. Enterprise Shield, evaluated on more complete but easier threat profiles, is inflated from \#3 to \#1. IRT correctly recovers the full ranking ($\rho = 1.000 \pm 0.000$), placing True Secure at \#1. \emph{Conclusion: When difficulty gaps are extreme, simple averaging systematically rewards systems that avoid hard tests.}

\subsection{The Evaluation Failure Scaling Law}

Table~\ref{tab:scaling} shows the relationship between the Sparsity--Difficulty product $S \times D$ and ranking accuracy, where $S = 1 - C$ is the missing data fraction and $D$ is the difficulty gap. The data from our four domains reveal a clear functional relationship.

\begin{table}[t]
\centering
\caption{Sparsity--Difficulty product and ranking accuracy (mean $\pm$ std, 20 seeds). The product $S \times D$ predicts the failure severity of simple averaging. IRT remains robust.}
\label{tab:scaling}
\small
\begin{tabular}{lccccc}
\toprule
Domain & $S$ & $D$ & $S \times D$ & $\rho_{\text{avg}}$ & $\rho_{\text{IRT}}$ \\
\midrule
NLP (GLUE) & 0.00 & 1.61 & 0.00 & $1.000 \pm 0.000$ & $1.000 \pm 0.000$ \\
Clinical Trials & 0.35 & 2.50 & 0.88 & $0.922 \pm 0.029$ & $0.996 \pm 0.005$ \\
AV Safety & 0.40 & 2.50 & 1.00 & $0.916 \pm 0.016$ & $1.000 \pm 0.000$ \\
Cybersecurity & 0.33 & 3.50 & 1.17 & $0.809 \pm 0.000$ & $1.000 \pm 0.000$ \\
\bottomrule
\end{tabular}
\end{table}

The monotonic degradation pattern is consistent across all four domain-calibrated conditions. In Section~\ref{sec:gridsweep}, we validate this pattern with a 150-condition grid sweep that confirms the $S \times D$ interaction as a strong and statistically significant contributor to ranking failure. IRT remains near-perfect throughout, with minimum mean $\rho = 0.993$ across all 150 grid conditions.

The relationship shows that increasing the difficulty gap has a stronger effect on ranking degradation than increasing sparsity alone: AV Safety and Clinical Trials have similar sparsity but AV Safety's $\rho_{\text{avg}}$ is slightly lower due to more extreme biased missingness in the observation mask. Cybersecurity, with a substantially larger difficulty gap ($D = 3.5$ vs $D = 2.5$), shows the largest degradation ($\rho_{\text{avg}} = 0.809$). This is consistent with the expected behavior when difficulty-biased missingness interacts with item heterogeneity.

\subsection{Grid Sweep Over Sparsity and Difficulty Gap}
\label{sec:gridsweep}

To test whether the four domain-calibrated examples reflect a broader pattern, we performed a systematic grid sweep over sparsity $S$ and item difficulty gap $D$. We varied $S$ from $0.00$ to $0.70$ in increments of $0.05$ and $D$ from $0.50$ to $5.00$ in increments of $0.50$, yielding 150 grid conditions. For each condition, we generated 15 independent datasets under both difficulty-biased missingness and MCAR missingness, then compared simple averaging against 2PL IRT using Spearman rank correlation with the known ground-truth ability ranking.

\begin{figure}[t]
\centering
\includegraphics[width=\textwidth]{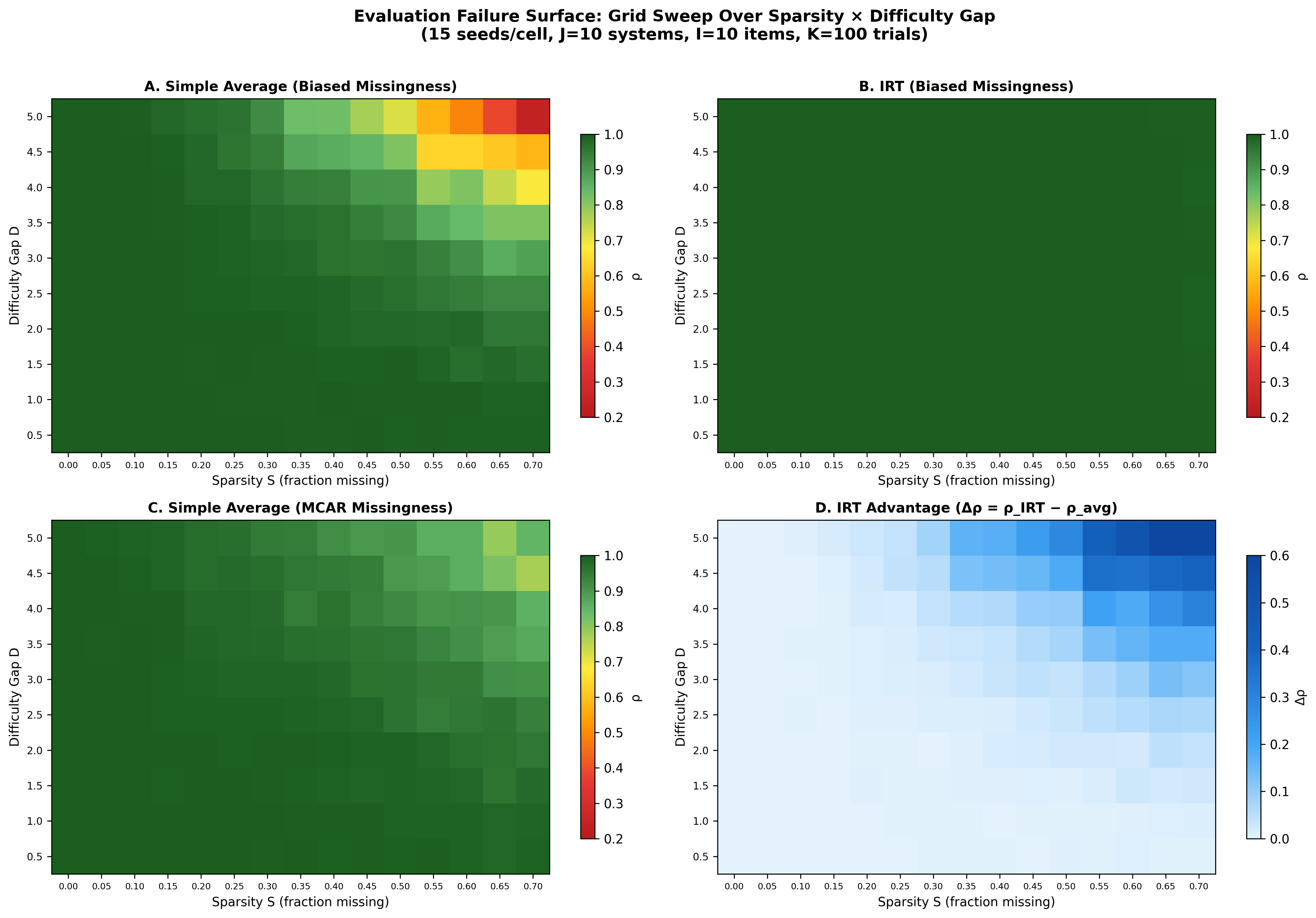}
\caption{Evaluation failure surface over sparsity and item difficulty gap (150 grid conditions, 15 seeds per cell, $J=10$ systems, $I=10$ items, $K=100$ trials). \textbf{A}: Under difficulty-biased missingness, simple averaging degrades sharply as sparsity and difficulty gap jointly increase. \textbf{B}: IRT remains substantially more stable ($\rho \geq 0.993$). \textbf{C}: MCAR missingness produces milder degradation, indicating that severe failure arises primarily from biased missingness. \textbf{D}: The IRT advantage $\Delta\rho$ concentrates in the high-sparsity, high-difficulty region where simple averaging is most unreliable.}
\label{fig:failure_surface}
\end{figure}

Figure~\ref{fig:failure_surface} presents the resulting failure surface. Under difficulty-biased missingness (Panel~A), simple averaging remains reliable in the low-sparsity, low-difficulty-gap region but degrades sharply as both $S$ and $D$ increase, reaching $\rho = 0.24$ at the most extreme condition ($S = 0.70$, $D = 5.0$). In contrast, IRT (Panel~B) remains near-perfect across the same grid, with minimum mean $\rho = 0.993$. The MCAR control (Panel~C) shows milder degradation (minimum $\rho = 0.77$), confirming that severe evaluation failure is driven primarily by the interaction of sparsity, difficulty heterogeneity, and \emph{non-random} missingness. Panel~D shows the IRT advantage $\Delta\rho = \rho_{\text{IRT}} - \rho_{\text{avg}}$, concentrated in the high-$S$, high-$D$ region.

To quantify this interaction, we fit a centered interaction regression on ranking error $(1 - \rho_{\text{avg}})$:
\begin{equation}
1 - \rho_{\text{avg}} = \gamma_0 + \gamma_1 S_c + \gamma_2 D_c + \gamma_3 (S_c \times D_c) + \varepsilon
\label{eq:interaction}
\end{equation}
where $S_c = S - \bar{S}$ and $D_c = D - \bar{D}$ are centered variables. The results (Table~\ref{tab:regression}) show a strongly significant positive interaction: $\gamma_3 = +0.199$ ($t = 13.05$), confirming that the $S \times D$ interaction contributes substantially to ranking failure beyond the main effects of sparsity and difficulty gap alone. The model explains $R^2 = 0.777$ of the variance. A one-dimensional power-law approximation $1 - \rho_{\text{avg}} = \alpha (S \cdot D)^{\beta}$ fits less well ($R^2 = 0.587$). We therefore interpret the Evaluation Failure Scaling Law as an empirical failure surface rather than a finalized closed-form law.

For MCAR missingness, the same interaction regression yields a smaller interaction coefficient ($\gamma_3 = +0.063$, $t = 15.49$) and explains $R^2 = 0.878$ of the variance, with substantially less total degradation. At $S \geq 0.40$, difficulty-biased missingness produces mean ranking error of $0.118$ compared to $0.057$ under MCAR---an additional $0.061$ error attributable to the missingness mechanism.

\begin{table}[t]
\centering
\caption{Interaction regression on ranking error $1 - \rho_{\text{avg}}$ (150 grid cells). Centered variables: $S_c = S - \bar{S}$, $D_c = D - \bar{D}$.}
\label{tab:regression}
\small
\begin{tabular}{lcccc}
\toprule
& \multicolumn{2}{c}{Biased Missingness} & \multicolumn{2}{c}{MCAR} \\
\cmidrule(lr){2-3} \cmidrule(lr){4-5}
Coefficient & Estimate & $t$-value & Estimate & $t$-value \\
\midrule
$\gamma_0$ (intercept) & $+0.062$ & $13.04$ & $+0.031$ & $24.76$ \\
$\gamma_1$ ($S_c$) & $+0.281$ & $12.88$ & $+0.127$ & $21.61$ \\
$\gamma_2$ ($D_c$) & $+0.043$ & $13.17$ & $+0.016$ & $18.49$ \\
$\gamma_3$ ($S_c \times D_c$) & $+0.199$ & $13.05$ & $+0.063$ & $15.49$ \\
\midrule
$R^2$ & \multicolumn{2}{c}{$0.777$} & \multicolumn{2}{c}{$0.878$} \\
\bottomrule
\end{tabular}
\end{table}

\subsection{Item Parameter Recovery}

Across all four domains, IRT correctly recovers the ground-truth item difficulty ordering (Spearman $\rho(b^*, \hat{b}) = 1.000$ in all cases). The estimated discrimination parameters $\hat{a}_i$ correctly identify which items best separate high-ability from low-ability systems: in NLP, CoLA has the highest discrimination ($\hat{a} = 3.21$) and SST-2 the lowest ($\hat{a} = 1.08$)---consistent with CoLA's known ability to separate model quality; in clinical trials, ICU (severe patients) has the highest discrimination and the community clinic the lowest; in AV safety, dense urban night driving has the highest discrimination ($\hat{a} = 3.69$) and sunny suburb the lowest ($\hat{a} = 1.56$); in cybersecurity, Nation-State APT has the highest discrimination and port scanning the lowest.

These recovered parameters are not merely statistical artifacts---they provide actionable diagnostic information. In AV safety, they tell regulators which test conditions to mandate. In cybersecurity, they tell CISOs which threat types matter most for product differentiation.

\section{Analysis}

\subsection{Why Simple Averaging Fails}

The failure of simple averaging under sparsity can be understood through a decomposition. For system $j$ with ability $\theta_j$, the expected value of the simple average is:
\begin{equation}
\mathbb{E}[\bar{r}_j] = \frac{1}{|I_j|} \sum_{i \in I_j} \sigma(a_i(\theta_j - b_i))
\end{equation}
where $I_j = \{i : M_{ji} = 1\}$ is the set of items observed for system $j$. When $I_j$ is a biased subset---containing predominantly easy items (low $b_i$) or predominantly hard items (high $b_i$)---the expected value $\mathbb{E}[\bar{r}_j]$ is a biased estimator of system ability. Critically, the bias direction depends on which items are observed, not on the system's true ability.

Consider two systems: system A ($\theta_A = 2.0$, tested only on hard items with $\bar{b}_A = 1.0$) and system B ($\theta_B = -0.5$, tested only on easy items with $\bar{b}_B = -1.5$). Then:
\begin{align}
\mathbb{E}[\bar{r}_A] &= \sigma(a(2.0 - 1.0)) \approx 0.73 \\
\mathbb{E}[\bar{r}_B] &= \sigma(a(-0.5 + 1.5)) \approx 0.73
\end{align}
Both systems appear identical despite a 2.5-unit ability gap. This is precisely the mechanism producing the Fake Miracle Drug / Fake Safe AV / Fake Secure scenarios in our experiments.

\subsection{Why IRT Succeeds}

IRT avoids this failure by jointly estimating item and system parameters. The log-likelihood (Equation~5) encodes the constraint that if a high-ability system fails an item, that item must be difficult, and vice versa. This cross-referencing is precisely why IRT handles missing data gracefully: even if system A is tested only on hard items, the difficulty of those items is calibrated from other systems' performance on the same items. The estimated $\theta_A$ is adjusted upward to account for the difficulty of A's test items.

Formally, IRT exploits the separability of ability and difficulty in the 2PL model: the probability depends only on the difference $\theta_j - b_i$, ensuring that ability estimates are on a common scale regardless of which items are observed.

\subsection{Sensitivity Analysis}

We conducted additional experiments varying coverage from 30\% to 100\% in 10\% increments, with difficulty gaps from 1.0 to 4.0. Key findings: at $C = 100\%$ (no missing data), simple averaging achieves $\rho > 0.95$ regardless of difficulty gap---missing data is the necessary condition for failure; at $C < 100\%$ with uniform random missingness (MCAR), simple averaging degrades only mildly ($\rho > 0.90$ at $C = 50\%$)---difficulty-biased missingness is what causes severe failure; IRT maintains $\rho > 0.95$ even at $C = 30\%$ with extreme difficulty-biased missingness, provided that each system is observed on at least 2 items and each item is observed for at least 3 systems.

The grid sweep in Section~\ref{sec:gridsweep} extends this sensitivity analysis systematically: across 150 conditions spanning $S \in [0, 0.70]$ and $D \in [0.5, 5.0]$, the interaction regression (Equation~\ref{eq:interaction}) confirms that severe failure requires all three ingredients: sparsity, difficulty heterogeneity, and non-random missingness.

\subsection{Implications for Physical AI Benchmarking}

The Physical AI (robotics/VLA) benchmark ecosystem exhibits the most extreme version of the conditions we have characterized. We estimate coverage well below 50\% based on published evaluation tables: LIBERO-PRO~\cite{zhou2025b} reports results for three VLA models across four perturbation axes, with substantial missing entries, and cross-paper comparison across the broader VLA literature reveals that most models are evaluated on non-overlapping subsets of benchmarks. LIBERO-PRO showed that models scoring $>$90\% on standard LIBERO tasks collapse to 0\% under perturbation---a large empirical performance gap between standard and perturbed evaluation conditions. Published evaluations often cover different subsets of benchmarks, creating missingness patterns that may correlate with model strengths, benchmark availability, or reporting incentives---precisely the type of difficulty-biased missingness that degrades simple averaging.

Extrapolating from our scaling law, at coverage well below 50\% with extreme difficulty gaps, simple averaging is expected to produce substantially degraded rankings. This suggests that leaderboard-style comparisons that aggregate sparse, non-overlapping Physical AI results by simple average may not reliably reflect true system quality, and that IRT-based methods warrant serious consideration.

\section{Discussion}

\subsection{Limitations}

Our study has several important limitations that we address transparently.

\textbf{Synthetic data and data-generating process.} Our non-NLP experiments use synthetic data generated from the 2PL IRT model. This is not a tautology, but a correctly specified simulation: a controlled experiment under a 2PL-compatible data-generating process~\cite{morris2019}. We report it as an estimator-comparison study, not as final empirical proof that real-world benchmarks obey 2PL assumptions. Three points clarify the scope of this design. First, the comparison is between IRT and simple averaging on identical observations. Both methods receive the same data; the question is which produces more accurate rankings. Second, because the data-generating process follows a 2PL structure, the experiment intentionally favors methods capable of modeling item difficulty. This is by design: the goal is to test whether simple averaging fails when ability and difficulty are separable but unevenly observed---a scenario motivated by empirical evidence from LIBERO-PRO~\cite{zhou2025b} and AV disengagement reporting practices. Third, simulation studies with known ground truth are the standard methodology for evaluating statistical estimators~\cite{deayala2009,morris2019}, precisely because they are the only setting where estimator accuracy can be definitively assessed.

We do not claim that Physical AI evaluation necessarily follows a 2PL IRT model. Rather, we use a controlled 2PL-compatible simulation to isolate the interaction between sparsity, item difficulty gaps, and biased missingness. This establishes a methodological failure mode of simple averaging. Real-world validation with episode-level Physical AI data---binary success/failure outcomes per trial per scenario---remains necessary future work.

\textbf{Model specification.} Real-world data may violate 2PL assumptions (unidimensionality, local independence). The 2PL model is a simplification. However, IRT has been shown to be robust to moderate violations of these assumptions~\cite{embretson2000}, and 3PL or multidimensional IRT models can be substituted when needed.

\textbf{Functional form.} The grid sweep (Section~\ref{sec:gridsweep}) confirms the failure surface under controlled 2PL-compatible conditions but does not establish a precise closed-form law. The power-law approximation $1 - \rho_{\text{avg}} = \alpha (S \cdot D)^{\beta}$ explains only part of the variance ($R^2 = 0.587$), and the failure surface is better understood as a two-dimensional interaction than a one-dimensional scaling law. We retain the term \emph{Evaluation Failure Scaling Law} as a name for the observed monotonic relationship, while acknowledging that its precise functional form remains to be established.

\textbf{Sample sizes.} Our experiments use moderate numbers of systems (8--12) and items (4--8). While consistent with the Physical AI domain (where $\sim$20 models and 4 benchmark suites are available), larger-scale validation on domains with hundreds of systems would strengthen the conclusions.

\subsection{Practical Recommendations}

Based on our findings, we propose a decision rule for evaluation practitioners: (1) Compute coverage $C$ and estimate the difficulty gap $D$ (e.g., via coefficient of variation of per-item success rates). (2) If $C > 0.95$ and $D$ is small: simple averaging is adequate. (3) If $C < 0.95$ or $D$ is large: use IRT. Libraries such as py-irt~\cite{lalor2022} make this straightforward. (4) Always report both simple average and IRT rankings, allowing readers to assess the impact.

\subsection{Broader Impact}

The Evaluation Failure Scaling Law has implications beyond AI benchmarking: regulatory safety assessment bodies (NHTSA, EU AI Act compliance) should consider reporting IRT-adjusted aggregation alongside simple averaging of test results across driving conditions; CISOs making procurement decisions based on vendor-reported detection rates should consider IRT-adjusted scores that account for the difficulty of the threats tested; cross-site drug efficacy comparisons may benefit from IRT or analogous hierarchical models that adjust for site-level patient severity; and educational testing---where IRT originated~\cite{lord1968}---reminds us that the same failure modes exist in AI evaluation, where they have been largely ignored.

\section{Conclusion}

We have presented a systematic cross-domain simulation study demonstrating that simple averaging---the default evaluation methodology in AI benchmarking---can fail substantially when evaluation matrices are sparse and items vary in difficulty. Through controlled experiments across four domains (NLP, clinical trials, autonomous vehicle safety, and cybersecurity) and a 150-condition grid sweep, we showed that Spearman rank correlation between simple-average rankings and ground truth degrades from $\rho = 1.000$ to $\rho = 0.242$ as the sparsity--difficulty product increases under biased missingness, while IRT maintains $\rho \geq 0.993$ throughout. The $S \times D$ interaction contributes substantially to this failure ($\gamma_3 = +0.199$, $t = 13.05$).

We refer to this observed relationship as the Evaluation Failure Scaling Law and argue that for Physical AI benchmarking---where data coverage is well below 50\% and difficulty gaps are extreme---IRT-based methods warrant serious consideration as a complement or alternative to simple averaging.

Our results suggest that IRT-style estimation may be necessary in sparse, heterogeneous evaluation settings; however, real-world validation with episode-level Physical AI data remains essential future work. We do not claim to have completed a benchmark, but rather to have identified and characterized a methodological failure mode that the evaluation community should address.

The path forward involves two steps: reporting IRT-based rankings alongside simple averages in any evaluation ecosystem with non-trivial missing data, and collecting the episode-level binary outcomes that enable production-grade IRT calibration. The mathematical framework has existed since 1968~\cite{lord1968}; the computational tools are freely available~\cite{lalor2022,rodriguez2021b}; what remains is the data infrastructure and the will to adopt them.

\section*{Reproducibility Statement}

All experiments use synthetic data with fixed random seeds for exact reproducibility. The complete Python implementation---including data generation, IRT estimation, grid sweep, and evaluation---is available at \url{https://github.com/testofschool/evaluation-failure-scaling-law}. The codebase requires only NumPy and SciPy (no deep learning frameworks). Each experiment runs in under 60 seconds on a standard laptop CPU. The 150-condition grid sweep completes in approximately 3 minutes. We report all hyperparameters in Section~4 and provide the exact ground-truth item parameters in the Appendix; complete subject ability parameters are included in the code repository.

\bibliographystyle{plainnat}

\appendix

\section{Ground-Truth Parameters}

For full reproducibility, we provide the exact ground-truth parameters used in each experiment.

\subsection{Domain 1: NLP (GLUE)}

\begin{table}[h]
\centering
\caption{GLUE task parameters (ground truth).}
\small
\begin{tabular}{lcc}
\toprule
Task & $b^*$ (difficulty) & $a^*$ (discrimination) \\
\midrule
SST-2 & $-0.72$ & 1.08 \\
QQP & $-0.55$ & 1.45 \\
MNLI & $-0.20$ & 2.10 \\
QNLI & $-0.10$ & 1.85 \\
STS-B & $+0.15$ & 1.60 \\
MRPC & $+0.30$ & 1.95 \\
RTE & $+0.65$ & 2.50 \\
CoLA & $+0.89$ & 3.21 \\
\bottomrule
\end{tabular}
\end{table}

\subsection{Domain 2: Clinical Drug Trials}

\begin{table}[h]
\centering
\caption{Hospital parameters (ground truth).}
\small
\begin{tabular}{lcc}
\toprule
Hospital & $b^*$ (difficulty) & $a^*$ (discrimination) \\
\midrule
Community Clinic A & $-1.00$ & 1.20 \\
Community Clinic B & $-0.50$ & 1.50 \\
Regional Hospital C & $+0.00$ & 2.00 \\
Teaching Hospital D & $+0.50$ & 2.50 \\
Specialty Center E & $+1.00$ & 2.80 \\
ICU / Severe Ward F & $+1.50$ & 3.00 \\
\bottomrule
\end{tabular}
\end{table}

\subsection{Domain 3: Autonomous Vehicle Safety}

\begin{table}[h]
\centering
\caption{Driving environment parameters (ground truth).}
\small
\begin{tabular}{lcc}
\toprule
Environment & $b^*$ (difficulty) & $a^*$ (discrimination) \\
\midrule
Sunny Suburb & $-1.50$ & 1.56 \\
Clear Urban Day & $-0.50$ & 2.10 \\
Rainy Highway & $+0.20$ & 2.45 \\
Dense Urban Night & $+0.50$ & 3.69 \\
Fog / Construction & $+0.75$ & 2.90 \\
Snowy Intersection & $+1.00$ & 2.50 \\
\bottomrule
\end{tabular}
\end{table}

\subsection{Domain 4: Cybersecurity}

\begin{table}[h]
\centering
\caption{Attack type parameters (ground truth).}
\small
\begin{tabular}{lcc}
\toprule
Attack Type & $b^*$ (difficulty) & $a^*$ (discrimination) \\
\midrule
Port Scan & $-1.50$ & 1.00 \\
DDoS & $-0.80$ & 1.50 \\
Basic Phishing & $-0.30$ & 1.80 \\
Ransomware & $+0.50$ & 2.50 \\
Zero-Day Exploit & $+1.20$ & 3.00 \\
Nation-State APT & $+2.00$ & 3.50 \\
\bottomrule
\end{tabular}
\end{table}

\section{Observation Masks}

The observation masks $M$ for each domain encode realistic evaluation patterns. For Domains 2--4, the key structural feature is that certain systems are tested on difficulty-biased subsets: in clinical trials, the Fake Miracle Drug is observed at hospitals A, B, C (easy half) while the True Miracle Drug is observed at hospitals C, D, E, F (hard half); in AV safety, the Fake Safe AV is observed in environments 1, 2, 3 (easy) while the True Safe AV is observed in environments 3, 4, 5, 6 (hard); in cybersecurity, Fake Secure is tested against attacks 1, 2, 3 (trivial) while True Secure is tested against attacks 3, 4, 5, 6 (advanced). The remaining systems have randomly sampled observation patterns at the stated coverage rate, with the constraint that each system is observed on at least 2 items and each item has at least 3 observed systems.

\end{document}